\documentclass{llncs}

\usepackage{graphicx,verbatim,color,epsfig,latexsym,amsmath}

\usepackage{algorithm}
\usepackage{clrscode}
\usepackage{xspace}

\fontfamily{ptm}
\newcommand{\mxsat}{{\scshape MaxSat}\xspace}

\newenvironment{Proof}
               {\smallskip{\noindent{\bf Proof:}}\normalfont\slshape}
               {\hfill\rule{2mm}{2mm}\\}  

\begin{document}

\def\baselinestretch{0.995}

\title{On Using Unsatisfiability for \\ Solving Maximum Satisfiability}

\author{Joao Marques-Silva \and Jordi Planes}

\institute{School of Electronics and Computer Science, University of
  Southampton, UK \\
\email{\{jpms,jp3\}@ecs.soton.ac.uk}
}

\maketitle

\begin{abstract}
Maximum Satisfiability (\mxsat) is a well-known optimization pro\-blem,
with several practical applications.
The most widely known \mxsat algorithms are ineffective at solving
hard problems instances from practical application domains.
Recent work proposed using efficient Boolean Satisfiability (SAT)
solvers for solving the \mxsat problem, based on identifying and
eliminating unsatisfiable subformulas. However, these algorithms do
not scale in practice.
This paper analyzes existing \mxsat algorithms based on unsatisfiable
subformula identification. Moreover, the paper proposes a number of
key optimizations to these \mxsat algorithms and a new alternative
algorithm. The proposed optimizations
and the new algorithm provide significant performance improvements on
\mxsat instances from practical applications.
Moreover, the efficiency of the new generation of
unsatisfiability-based \mxsat solvers becomes effectively indexed to
the ability of modern SAT solvers to proving unsatisfiability and
identifying unsatisfiable subformulas.
\end{abstract}

\section{Introduction}
\label{sec:intro}

The problem of Maximum Satisfiability~(\mxsat) consists of identifying
the largest number of clauses in a CNF formula that can be satisfied.
Variations of the \mxsat include partial \mxsat and weighted \mxsat.
For partial \mxsat some clauses (i.e.~the hard clauses) must be
satisfied whereas others (i.e.~the soft clauses) may not be
satisfied. For weighted \mxsat, each clause has a given weight, and
the objective is to maximize the sum of the weights of satisfied
clauses.

The \mxsat problem and its variations find a number of relevant
practical applications, including design debugging of embedded
systems~\cite{kas-fmcad07} and FPGA routing~\cite{kas-tcad03}.
Unfortunately, the techniques that have proved to be extremely
effective in Boolean Satisfiability~(SAT) cannot be applied directly
to \mxsat~\cite{manya-aij07,heras-aij07}. As a result, most of the
existing algorithms~\cite{HLO07,LMP06,LMP07,FM06} implement only a
restricted  number of techniques, emphasizing bound computation and/or
dedicated inference techniques. Despite the extensive research work in
this area, existing \mxsat techniques and algorithms do not scale for
large problem instances from practical applications.

Recent work~\cite{FM06} proposed alternative approaches, that build
on the existence of effective SAT solvers for identifying
unsatisfiable subformulas, and so can indirectly exploit existing
effective SAT techniques~\cite{jpms-iccad96,malik-dac01,een-sat03}.
However, even though modern SAT solvers are effective at proving
unsatisfiability and generating unsatisfiable subformulas, the
algorithms described in~\cite{FM06} are in general ineffective for
\mxsat, and so this work focused on partial \mxsat with a reduced
number of soft clauses.

This paper reviews previous \mxsat algorithms based on identifying
unsatisfiable subformulas for \mxsat, proposes key optimizations to 
one of these algorithms~\cite{FM06}, and develops a new algorithm
also based on identifying unsatisfiable subformulas. Experimental
results, obtained on a wide range of practical problem instances,
show that the new \mxsat algorithms can be orders of magnitude more
efficient than the original algorithms~\cite{FM06}, being in general
consistently more efficient than previous \mxsat solvers on instances
obtained from practical applications.

The paper is organized as follows. Section~\ref{sec:defs} briefly
introduces the \mxsat problem and existing algorithms.
Afterwards, Section~\ref{sec:msu} reviews \mxsat algorithms based on
unsatisfiable subformula identification~\cite{FM06}.
Section~\ref{sec:nmsu1} proposes optimizations to these algorithms,
and Section~\ref{sec:nmsu3} proposes a new \mxsat algorithm.
Experimental results on a large sample of problem instances, obtained
from a number of practical applications, are analyzed in
Section~\ref{sec:res}. The paper concludes in Section~\ref{sec:conc}.


\section{Preliminaries}
\label{sec:defs}

This section provides definitions and background knowledge for the
\mxsat problem. Due to space constraints, familiarity with SAT and
related topics is assumed~\cite{jpms-iccad96,een-sat03}.

The maximum satisfiability (\mxsat) problem can be stated as
follows. Given an instance of SAT represented in Conjunctive Normal
Form (CNF), compute an assignment to the variables that maximizes the
number of satisfied clauses. 
Variations of the \mxsat problem include the partial \mxsat and the
weighted \mxsat problem.
In the partial MaxSAT problem some clauses (i.e.~the {\em hard}
clauses) must be satisfied, whereas others (i.e.~the {\em soft}
clauses) may not be satisfied. In the weighted MaxSAT problem, each
clause has a given weight, and the objective is to maximize the sum of
the weights of satisfied clauses.

During the last decade there has been a growing interest on studying
\mxsat, motivated by an increasing number of practical applications,
including
scheduling, 
routing, 
bioinformatics, 
and design automation~\cite{kas-tcad03,kas-fmcad07}.
Despite the clear relationship with the SAT problem, most modern SAT
techniques cannot be applied directly to the \mxsat
problem~\cite{manya-aij07,heras-aij07}. As a result, most \mxsat
algorithms are built on top of the standard DPLL~\cite{davis-cacm62}
algorithm, and so do not scale for industrial problem
instances~\cite{HLO07,LMP06,LMP07,FM06}.
The most often used approach for \mxsat (e.g.~most of the solvers in
the \mxsat competition~\cite{maxsatcomp}) is based on a Branch and
Bound algorithm, emphasizing the computation of a lower bound and the
application of inference rules that simplify the
instance~\cite{HLO07,LMP06,LMP07}.
Results from the \mxsat competition~\cite{maxsatcomp} indicate that
solvers based on Branch and Bound with additional inference rules are
currently the most efficient \mxsat solvers, outperforming all other
existing approaches.

One alternative approach for solving the \mxsat problem is to use
Pseudo-Boolean Optimization (PBO) (e.g.~\cite{kas-sat05}). The PBO
approach for \mxsat consists of adding a new ({\em blocking)} variable
to each clause. The blocking variable $b_i$ for clause $\omega_i$
allows satisfying clause $\omega_i$ independently of other assignments
to the problem variables. The resulting PBO formulation includes a
cost function, aiming the minimization of the number of blocking
variables assigned value 1.
Clearly, the solution of the \mxsat problem is obtained by subtracting
from the number of clauses the solution of the PBO problem.

Despite its simplicity, the PBO formulation does not scale for
industrial problems, since the large number of clauses results in
a large number of blocking variables, and corresponding larger search
space. Observe that, for most instances from practical applications,
the number of clauses largely exceeds the number of variables.
For the resulting PBO problem, the number of variables equals the sum
of the number of variables and clauses in the original SAT
problem. Hence, the resulting instance of PBO has a much larger search
space than the original instance of SAT.

Besides the PBO model, a number of alternative algorithms exist for
\mxsat. Examples include, {\tt OPT-SAT}~\cite{giunchiglia-ecai06} and
{\tt sub-SAT}~\cite{kas-tcad03}. {\tt OPT-SAT} imposes an ordering on
the Boolean variables on an existing SAT solver. Experimental results
for \mxsat indicate that this approach is slower than a state-of-art
PBO solver, e.g.~{\tt minisat+}~\cite{een-jsat06}, and so it is
unlikely to scale for industrial problems. On the other hand, {\tt
  sub-SAT} solves a relaxed version of the original problem, hence the
exact \mxsat solution may not be computed. Moreover, the experimental
comparison in~\cite{FM06} suggests that {\tt sub-SAT} is not
competitive with unsatisfiability-based \mxsat algorithms.
Other approaches have been proposed~\cite{mazure-ijcai07}, that are
based on the relation of minimally unsatisfiable subformulas and
maximally satisfiable subformulas~\cite{kas-sat05,mazure-ijcai07}.
However, these approaches are based on enumeration of maximally
satisfiable subformulas, and so do not scale for instances with a
large number of unsatisfiable subformulas. As a result, for most
instances only approximate results can be obtained.
More recently, an alternative approximate approach to \mxsat has been
proposed~\cite{kas-fmcad07}. The motivation for this alternative
approach is the potential application of \mxsat in design debugging,
and the fact that existing \mxsat approaches do not scale for
industrial problem instances. However, this approach is unable to
compute exact solutions to the \mxsat problem.

The next section addresses \mxsat algorithms that use the
identification of unsatisfiable subformulas.
Modern SAT solvers can be instructed to generate a resolution
refutation for unsatisfiable formulas~\cite{zhang-date03}. The
resolution proof is usually represented as a proof trace, which
summarizes the resolution steps used for creating each clause learnt
by the SAT solver. 
Besides resolution refutations, proof traces allow identifying
unsatisfiable subformulas, which serve as the source for the
resolution refutation. A simple iterative procedure allows generating
minimal unsatisfiable subformulas (MUS) from computed unsatisfiable
sub-formulas~\cite{zhang-date03}.
All modern conflict-driven clause learning (CDCL) SAT solvers can be
easily adapted to generate proof traces, and indirectly, unsatisfiable
sub-formulas.


\section{Unsatisfiability-Based MaxSat Algorithms}
\label{sec:msu}

As mentioned in the previous section, one of the major drawbacks of
the PBO model for \mxsat is the large number of blocking variables
that must be considered. The ability to reduce the number of required
blocking variables is expected to improve significantly the ability of
SAT/PBO based solvers for tackling instances of \mxsat.
Moreover, any solution to the \mxsat problem will be unable to satisfy
clauses that {\em must} be part of an unsatisfiable subformula.
Consequently, one approach for reducing the number of blocking
variables is to associate blocking variables only with clauses that
are part of unsatisfiable subformulas. However, it is not simple to
identify all clauses that are part of unsatisfiable subformulas. One
alternative is the identification and relaxation of unsatisfiable
subformulas.

This section describes the unsatisfiability-based \mxsat algorithm
described in~\cite{FM06}. In what follows this algorithm is referred
to as {\tt msu1} (Fu\&Malik's \mxsat algorithm based on unsatisfiable
subformulas). It should be observed that the original algorithm was
proposed for partial \mxsat, but the modifications for the plain
\mxsat problem are straightforward. 

Algorithm~\ref{alg:msu1} summarizes Fu\&Malik's~\cite{FM06} \mxsat
algorithm.
The algorithm iteratively finds unsatisfiable cores
(line~\ref{li:msu:satcall}), adds new blocking variables to the
non-auxiliary clauses in the unsatisfiable core
(line~\ref{li:msu:newbv}), and requires that exactly one of the new
blocking  variables must be assigned value 1
(line~\ref{li:msu:card}). This constraint is referred to as the {\em
  One-Hot} constraint in~\cite{FM06}.
The algorithm terminates whenever the CNF formula is satisfiable, and
the number of assigned blocking variables is used for computing the 
solution to the \mxsat problem instance.

The clauses used for implementing the {\em One-Hot} constraint are
declared auxiliary; all other clauses are non-auxiliary. Observe that
each non-auxiliary clause may receive more than one blocking variable,
and the total number of blocking variables a clause receives
corresponds to the number of times the clause is part of an
unsatisfiable core. As suggested earlier in this section, by focusing
on identification and relaxation (with blocking variables) of
unsatisfiable sub-formulas, {\tt msu1} and the other algorithms
described later attempt to reduce the number of blocking variables
that is necessary to use while solving the \mxsat problem.

%
\begin{algorithm}[t]
{\small
\begin{codebox}
\Procname{$\proc{msu1}(\varphi)$}
\li \Comment Clauses of CNF formula $\varphi$ are the {\em initial} clauses
\li \Comment Clauses in $\varphi$ are tagged non-auxiliary
\li $\varphi_{W} \gets \varphi$ \RComment Working formula, initially
set to $\varphi$
\li \While \kw{true}\label{li:msu:init-loop}
\li   \Do $(\textnormal{st}, \varphi_C) \gets \textsc{SAT}(\varphi_{W})$\label{li:msu:satcall}
\li   \Comment{$\varphi_C$ is an unsat core if $\varphi_W$ is unsat}
\li   \If $\textnormal{st} = \kw{UNSAT}$
\li     \Then\label{li:msu:unsat}
           $BV \gets \emptyset$
\li        \For\kw{each} $\omega\in\varphi_C$
\li           \Do
              \If $\omega$ is not auxiliary
\li              \Then 
                    $b$ is a new blocking variable
\li                 $\omega_B \gets \omega \cup \{ b \}$ \label{li:msu:newbv} \RComment $\omega_B$ is tagged non-auxiliary
\li                 $\varphi_W \gets\varphi_W - \{ \omega \}\cup \{ \omega_B \}$
\li                 $BV \gets BV \cup \{ b \}$ \End\End
\li        $\varphi_B \gets \textnormal{CNF}(\sum_{b\in BV} b = 1)$\label{li:msu:card}
\RComment {\em One-Hot} constraint in~\cite{FM06}
\li        $\varphi_w \gets \varphi_W \cup \varphi_B$ \RComment Clauses in $\varphi_B$ are tagged auxiliary
\li     \Else\label{li:msu:sat} \Comment{Solution to \mxsat problem}
\li        $\nu\gets |\,\textnormal{blocking variables w/ value 1}\,|$
\li        \Return $|\varphi| - \nu$ \End\label{li:msu:sol}
\End
\end{codebox}
}
\caption{The \mxsat algorithm of Fu\&Malik}
\label{alg:msu1}
\end{algorithm}

%
A proof of correctness of algorithm {\tt msu1} is given
in~\cite{FM06}. However,~\cite{FM06} does not address important 
properties of the algorithm, including the number of blocking
variables that must be used in the worst case, or the worst-case
number of iterations of the algorithm.
This section establishes some of these properties. In what follows,
$n$ denotes the number of variables and $m$ denotes the number of
clauses.

\begin{proposition}
\label{prop:mbs}
During the execution of Algorithm~\ref{alg:msu1}, non-auxiliary
clauses can have multiple blocking variables.
\end{proposition}

\begin{Proof}
Consider the following example CNF formula:
\begin{equation*}
\begin{array}{l}
(x_1)\wedge(\neg x_1\vee\neg y_1)\wedge(y_1)\wedge(\neg x_1\vee\neg z_1)\wedge(\neg y_1\vee\neg z_1) \\
(x_2)\wedge(\neg x_2\vee\neg y_2)\wedge(y_2)\wedge(\neg x_2\vee\neg z_1)\wedge(\neg y_2\vee\neg z_1) \\
(z_1\vee z_2)\wedge(z_1\vee\neg z_2)\\
\end{array}
\end{equation*}
One possible execution of the algorithm follows.
Identify core $(x_1)\wedge(\neg x_1\vee\neg y_1)\wedge(y_1)$. Add
blocking clauses, respectively $b_1, b_2, b_3$, and require
$b_1+b_2+b_3 = 1$.
Identify core $(x_2)\wedge(\neg x_2\vee\neg y_2)\wedge(y_2)$. Add
blocking clauses, respectively $b_4, b_5, b_6$, and require
$b_4+b_5+b_6 = 1$.
Identify core $(x_1\vee b_1)\wedge(y_1\vee b_3)\wedge(\neg x_1\vee
\neg z_1)\wedge(\neg y_1\vee\neg z_1)\wedge(z_1\vee z_2)\wedge(z_1
\vee\neg z_2)\wedge\varphi_e$, where $\varphi_e$ denotes clauses from
encoding $b_1+b_2+b_3=1$ in CNF. Add blocking clauses to non-auxiliary
clauses, respectively $b_7, b_8, b_9, b_{10}, b_{11}, b_{12}$, and
require $b_7 + b_8 + b_9 + b_{10} + b_{11} + b_{12} = 1$.
At this stage, some of the non-auxiliary clauses have two blocking
variables, e.g.~$b_1$ and $b_7$ are associated with $(x_1)$.
\end{Proof}

\begin{proposition}
\label{prop:nbs}
During the execution of Algorithm~\ref{alg:msu1}, for iteration $j$,
exactly $j-1$ blocking variables must be assigned value 1, or the
formula is unsatisfiable.
\end{proposition}

\begin{Proof}
Observe that each iteration adds a constraint requiring the sum of a
set of new blocking variables to be equal to 1. Hence, at iteration
$j$, either $j-1$ blocking variables are assigned value 1, or the
formula is unsatisfiable.
\end{Proof}

\begin{proposition}
\label{prop:nbcs}
During the execution of Algorithm~\ref{alg:msu1}, if $\varphi_W$ is
satisfiable, at most 1 of the blocking variables associated with a
given clause can be assigned value 1.
\end{proposition}

\begin{Proof}
Each blocking variable is associated with a clause as the result of
identifying an unsatisfiable core.
Consider clause $\omega_i$ that is part of two cores $c_1$ and
$c_2$, each adding to $\omega_i$ a blocking variable, respectively
$b_{i,1}$ and $b_{i,2}$.
Assume that the formula could be satisfied such that $\omega_i$ would
have the two blocking variables $b_{i,1}$ and $b_{i,2}$ assigned 
value~1.
This would imply that both cores $c_1$ and $c_2$ could be deactivated
by blocking clause~$\omega_i$. But this would also imply that the
second core $c_2$ could not have been identified, since assigning
$b_{i,1}$ would deactivate core $c_2$; a contradiction.
\end{Proof}

This result allows deriving an upper bound on the number of iterations
of Algorithm~\ref{alg:msu1}.

\begin{proposition}
\label{prop:iter}
The number of iterations of Algorithm~\ref{alg:msu1} is ${\cal O}(m)$.
\end{proposition}

\begin{Proof}
Immediate from Propositions~\ref{prop:nbs} and~\ref{prop:nbcs}. At
each iteration $j$, $j-1$ blocking variables must be assigned value
1. Moreover, none of these blocking variables can be from the same
clause. Hence, at iteration $m+1$ all clauses must be satisfied by
assigning a blocking variable to 1.
Hence, the number of iterations of Algorithm~\ref{alg:msu1} is ${\cal
  O}(m)$.
\end{Proof}

It should be observed that the algorithm will {\em never} execute
$m+1$ steps. Indeed, for arbitrary CNF formulas, at least half of the
clauses can be trivially satisfied~\cite{johnson-stoc73},
and so the number of iterations never exceeds $\frac{m}{2}+1$.
Moreover, the upper bound on the number of iterations serves for
computing an upper bound on the total number of blocking variables.

\begin{proposition}
\label{prop:mxnbv}
During the execution of Algorithm~\ref{alg:msu1}, the number of
blocking variables is ${\cal O}(m^2)$ in the worst case.
\end{proposition}

\begin{Proof}
From Proposition~\ref{prop:iter} the number of iterations is
${\cal O}(m)$. In each iteration, each unsatisfiable core can have at
most $m$ clauses (i.e.~the number of original clauses). Hence the
result follows.
\end{Proof}

The previous result provides an upper bound on the number of blocking
variables. A tight lower bound is not known, even though a trivial
lower bound is $\Omega(m)$.


\section{Optimizing Unsatisfiability-Based \mxsat Algorithms}
\label{sec:nmsu1}

This section proposes improvements to Fu\&Malik's \mxsat
algorithm~\cite{FM06} described in the previous Section. The resulting
algorithm is referred to as {\tt msu2}.

\subsection{Encoding Cardinality Constraints}

The {\em one-hot} constraint used in {\tt msu1}~\cite{FM06}
corresponds to the well-known pairwise encoding for Equals~1
constraints~\cite{gent-ecai02}, i.e. cardinality constraints of the
form $\sum_{i=1}^{r} b_i = 1$. Usually, Equals~1 constraints 
are encoded with two constraints, one AtMost~1 constraint
(i.e.~$\sum_{i=1}^{r} b_i \le 1$) and one AtLeast~1 constraint
(i.e.~$\sum_{i=1}^{r} b_i \ge 1$).
It is also well-known that the pairwise encoding requires
$\frac{r\,(r-1)}{2}+1$ clauses, one clause for the AtLeast~1
constraint, and $\frac{r\,(r-1)}{2}$ binary clauses for the 
AtMost~1 constraint. Hence, the quadratic number of clauses results
from encoding the AtMost~1 constraint.
For large $r$, as is often the case for the \mxsat problem, the
pairwise encoding can require a prohibitively large number of clauses.
For example, for an unsatisfiable core with 10,000 clauses, the
resulting AtMost~1 constraint is encoded with 49,995,000 binary
clauses. For practical applications, unsatisfiable cores are likely to
exceed 10,000 clauses. As shown in Section~\ref{sec:res}, in many
cases, the pairwise encoding of an AtMost~1 constraint exhausts the
available physical memory resources.

There are a number of alternatives to the pairwise
encoding of AtMost~1
constraints~\cite{warners-ipl98,gent-ecai02,sinz-cp05,een-jsat06}, all
of which are linear in the number of variables in the constraint.
These encodings can either use sequential counters, sorters, or binary
decision diagrams (BDDs).
One simple alternative is to use BDDs for encoding a cardinality
constraint. A Boolean circuit is extracted from the BDD
representation, which can then be converted to CNF using Tseitin's
encoding~\cite{tseitin68}.
In most cases, the encoding takes into account the polarity
optimizations of Plaisted and Greenbaum~\cite{plaisted86,een-jsat06}
when generating the CNF formula. For the AtMost~1 constraint, the
BDD-based encoding of a cardinality constraint is linear in
$n$~\cite{een-jsat06}.
For the results in Section~\ref{sec:res}, the most significant
performance gains are obtained from using a BDD-based encoding for
AtMost~1 constraints, using Tseitin's encoding and
Plaisted\&Greenbaum's polarity optimizations.

One final remark is that Fu\&Malik's algorithm will also work if only 
AtMost~1 constraints are used instead of Equals~1 constraints. This
allows saving one (possibly quite large) clause in each iteration of
the algorithm.

\subsection{Blocking Variables Associated with each Clause}
\label{ssec:bvconstr}

Another potential drawback of Fu\&Malik's algorithm is that there
can be several blocking variables associated with a given clause (see
the analysis of Algorithm~\ref{alg:msu1}, including
Propositions~\ref{prop:mbs} and~\ref{prop:mxnbv}). 
Each time a clause $\omega$ is part of a computed unsatisfiable core,
a {\em new} blocking variable is added to $\omega$. Observe that
correctness of the algorithm requires that more than one blocking
variable may be associated with each clause. On the other hand,
despite the potentially large (but at most linear in $m$) number of
blocking variables associated with each clause, {\em at most} one of
these additional blocking variables can be used for actually
preventing the clause from participating in an unsatisfiable core (see
Proposition~\ref{prop:nbcs}).

A simple improvement for pruning the search space associated with
blocking variables is to require that {\em at most one} of the
blocking variables associated with a given clause $\omega$ can be
assigned value 1. For a given clause $\omega_i$, let $b_{i,j}$ be the
blocking variables associated with $\omega_i$. As a result, the
condition that at most 1 of the blocking variables of $\omega_i$ is
assigned value 1 is given by:
\begin{equation}
\sum_{j} b_{i,j} \le 1
\end{equation}

In general, the previous condition is useful when
Algorithm~\ref{alg:msu1} must execute a large number of iterations,
and many clauses are involved in a significant number of unsatisfiable
cores.

\begin{example}
Consider the example given in the proof of
Proposition~\ref{prop:mbs}. In the third iteration of the algorithm,
the first clause $(x_1)$ has been modified to $(x_1\vee b_1\vee
b_7)$. As a result, the CNF encoding of the additional constraint
$b_1+b_7\le 1$ can be added to the CNF formula. Since this is an
AtMost~1 constraint, the encoding proposed in the previous section
can also be used.
\end{example}

\section{A New Unsatisfiability-Based \mxsat Algorithm}
\label{sec:nmsu3}

This section proposes a new alternative algorithm for \mxsat. Compared
to the algorithms described in the previous sections, {\tt msu1} and
{\tt msu2}, the new algorithm guarantees that {\em at most 1} blocking
variable is associated with each clause. As a result, the worst case
number of blocking variables that can be used is $m$. Moreover, during
a first phase, the new algorithm extracts identified cores, whereas in
a second phase the algorithm addresses the problem of computing the
number of blocking variables that must be assigned value 1. The
objective of the first phase is to simplify identification of disjoint
unsatisfiable cores.

\begin{algorithm}[t]
{\small
\begin{codebox}
\Procname{$\proc{msu3}(\varphi)$}
\li $\varphi_{W} \gets \varphi$ \RComment Working formula, initially
set to $\varphi$
\li $\textnormal{UC} \gets\emptyset$
\li \While \kw{true}\label{li:msu3:first-loop}\RComment{Phase 1: Identify disjoint cores}
\li   \Do $(\textnormal{st}, \varphi_C) \gets \textsc{SAT}(\varphi_{W})$
      \RComment{$\varphi_C$ is an unsat core if $\varphi_W$ is unsat}
\li   \If $\textnormal{st} = \kw{UNSAT}$
\li     \Then\label{li:msu3:1:unsat}
           $\varphi_W \gets \varphi_W - \varphi_C$
\li        $\textnormal{UC}\gets\textnormal{UC}\cup\{\varphi_C\}$
\li     \Else\label{li:msu3:1:sat} \kw{break} \RComment{Move to 2nd loop}\label{li:msu3:1stloop-end}
\End\End
\li $\textnormal{BV}\gets\emptyset$\label{li:msu3:bv-start}
\li \For\kw{each} $\varphi_C\in\textnormal{UC}$\RComment{Add blocking variables}
\li    \Do \For\kw{each} $\omega\in\varphi_C$
\li       \Do 
             $b$ is a new blocking variable
\li          $BV \gets BV \cup \{ b \}$
\li          $\varphi_W\gets\varphi_W\cup\{ \omega\cup\{b\} \}$\label{li:msu3:bv-end}
          \End
       \End
\li $\lambda = |\textnormal{UC}|$\RComment{Lower bound on true blocking variables}
\li $\varphi_B \gets \textnormal{CNF}(\sum_{b\in BV} b=\lambda)$
\li $\varphi_W \gets \varphi_W \cup \varphi_B$\RComment{Current
  cardinality constraint}
\li \While \kw{true}\label{li:msu3:init-loop}\RComment{Phase 2:
  Increment lower bound $\lambda$}
\li   \Do $(\textnormal{st}, \varphi_C) \gets \textsc{SAT}(\varphi_{W})$
      \RComment{$\varphi_C$ is an unsat core if $\varphi_W$ is unsat}
\li   \If $\textnormal{st} = \kw{UNSAT}$
\li     \Then\label{li:msu3:unsat}
           $\lambda\gets\lambda + 1$
\li        \For\kw{each} $\omega\in\varphi_C$
\li           \Do
              \If $\omega$ has no blocking variable
\li              \Then 
                    $b$ is new blocking variable
\li                 $\omega_B \gets \omega \cup \{ b \}$ \RComment $\omega_B$ is tagged non-auxiliary
\li                 $\varphi_W \gets\varphi_W - \{ \omega \}\cup \{ \omega_B \}$
\li                 $BV \gets BV \cup \{ b \}$ \End\End
\li        $\varphi_W\gets\varphi_W-\varphi_B$
\li        $\varphi_B \gets \textnormal{CNF}(\sum_{b\in BV} b = \lambda)$
\RComment{New cardinality constraint}
\li        $\varphi_W \gets \varphi_W \cup \varphi_B$ \RComment Clauses in $\varphi_B$ are tagged auxiliary
\li     \Else\label{li:msu3:sat} 
           \Return $|\varphi| - \lambda$\label{li:msu3:sol}\RComment{Solution to \mxsat problem}
\End
\end{codebox}
}
\caption{A new \mxsat algorithm}
\label{alg:msu3}
\end{algorithm}

Algorithm~\ref{alg:msu3} shows the new \mxsat algorithm. 
The first phase of the algorithm is shown in
lines~\ref{li:msu3:first-loop} to~\ref{li:msu3:1stloop-end}. During
this phase disjoint cores are identified and removed from the
formula.
The first set of blocking variables are associated with each clause in
an unsatisfiable core in lines~\ref{li:msu3:bv-start}
to~\ref{li:msu3:bv-end}.
The second phase of the algorithm is shown in
lines~\ref{li:msu3:init-loop} to~\ref{li:msu3:sol}. During this phase
the lower bound on the number of blocking variables assigned value 1
is iteratively increased until the CNF formula becomes satisfiable.
For each identified unsatisfied core, a unique blocking variable is
associated with non-auxiliary clauses that do not have a blocking
variable.
The cardinality constraint $\sum b_i = k$ is encoded with one
AtLeast~$k$ ($\sum b_i\le k$) and one AtMost~$k$ ($\sum b_i\ge k$) 
constraints. As with {\tt msu2}, these constraints are represented
with BDDs and converted to CNF using Tseitin's
transformation~\cite{tseitin68} and including the polarity
optimizations of Plaisted and
Greenbaum~\cite{plaisted86,een-jsat06}. In this case the size of the
representation if ${\cal O}(r\cdot k)$, where $r$ is the number of
variables~\cite{een-jsat06} and $k$ is the cardinality constraint
bound.

Despite {\tt msu3} guaranteeing that the number of blocking variables
never exceeds $m$, there are a few potential drawbacks. The AtLeast
$k$ and AtMost~$k$ cardinality constraints used by {\tt msu3} are
significantly more complex to encode than the simple AtMost~1
constraint used by {\tt msu1} and {\tt msu2}. As a result, {\tt msu3}
is expected to perform better when the \mxsat solution is not far from
the total number of clauses.

As mentioned earlier for {\tt msu1}, Algorithm~\ref{alg:msu3} can use
AtMost~$k$ cardinality constraints instead of Equals~$k$ constraints. 
Finally, algorithm {\tt msu2} also allows evaluating whether two
phases can be useful for solving \mxsat. Clearly, the algorithm could
easily be modified to also use only one phase.


\section{Experimental Evaluation}
\label{sec:res}

This section evaluates a number of \mxsat solvers on industrial test
cases. Most instances are obtained from unsatisfiable industrial
instances used in past SAT competitions~\cite{satcomp} or available
from SATLIB~\cite{satlib}.
The classes of instances considered were the following:
\begin{enumerate}
\item Bounded model checking sintances from
  IBM~\cite{zarpas-sat05}. The problem instances were restricted to
  unsatisfiable instances, up to 35 computation steps, for a total of
  252.
\item Instances from the parametrized pipelined-processor verification
  problem~\cite{manolios-charme05}. The problem instances were
  restricted to the smallest 58 instances.
\item Verification of out-of-order microprocessors, from
  UCLID~\cite{bryant-fmcad02}. 31 unsatisfiable instances were
  considered.
\item Circuit testing instances~\cite{satlib}. 228 unsatisfiable
  instances were considered.
\item Automotive product configuration~\cite{sinz-aiedam03}. 84
  unsatisfiable instances were considered.
\end{enumerate}

In addition, instances from design debugging~\cite{kas-fmcad07} (29
unsatisfiable instances) and FPGA routing~\cite{kas-tcad03} (16
unsatisfiable instances) were also considered. These \mxsat instances
are known to be difficult, and most have no known \mxsat solutions.
As a result, the total number of problem instances used in the
experiments was 698.

The \mxsat solvers considered were the following: the best performing 
solver in the \mxsat 2007 evaluation~\cite{maxsatcomp},
{\tt maxsatz}~\cite{LMP06,LMP07}, a PBO formulation of the \mxsat
problem solved with {\tt minisat+}, one of the best performing PBO
solvers~\cite{een-jsat06,pbocomp}, an implementation of the algorithm
based on identification of unsatisfiable cores ({\tt
  msu1})~\cite{FM06}, {\tt msu1} with the improvements proposed in
Section~\ref{sec:nmsu1} ({\tt msu2}), and the new \mxsat algorithm
described in Section~\ref{sec:nmsu3} ({\tt msu3}).
{\tt msu1}, {\tt msu2} and {\tt msu3} are built on top of the same
unsatisfiable core extractor, implemented with {\tt
  minisat}~1.14~\cite{een-sat03}.

Other alternative \mxsat algorithms were not
considered~\cite{giunchiglia-ecai06,kas-tcad03,mazure-ijcai07,kas-sat05}.
Existing results for {\tt OPT-SAT} indicate that it is not competitive
with the PBO model solved with {\tt minisat+}. On the other hand, both
{\tt sub-SAT}~\cite{kas-tcad03} and {\tt HYCAM}~\cite{mazure-ijcai07}
only compute approximate solutions. Moreover, results
from~\cite{mazure-ijcai07} also show that existing approaches based on
enumerating all minimally unsatisfiable subformulas~\cite{kas-sat05}
are not competitive.

With respect to the PBO model, {\tt minisat+} was configured to use
sorters for the cost function. The reason for using sorters is that
for many problem instances the use of BDDs would exhaust the available
physical memory.

The results for all \mxsat solvers on all problem instances were
obtained on a Linux server running RHE Linux, with a Xeon 5160 3.0 GHz
dual-core processor. For the experiments, the available physical
memory of the server was 2 GByte. The time limit was set to 1000
seconds per instance.

\begin{figure}[t]
\begin{center}
\begin{picture}(0,0)%
\includegraphics{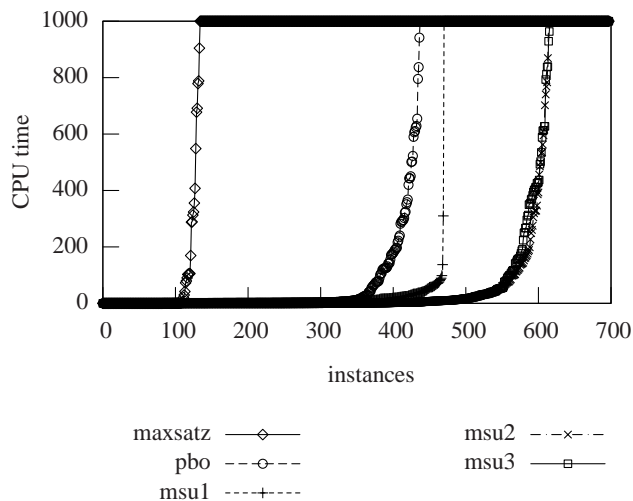}%
\end{picture}%
\begingroup
\setlength{\unitlength}{0.0200bp}%
\begin{picture}(12600,9720)(0,0)%
\put(1925,3850){\makebox(0,0)[r]{\strut{} 0}}%
\put(1925,4914){\makebox(0,0)[r]{\strut{} 200}}%
\put(1925,5978){\makebox(0,0)[r]{\strut{} 400}}%
\put(1925,7042){\makebox(0,0)[r]{\strut{} 600}}%
\put(1925,8106){\makebox(0,0)[r]{\strut{} 800}}%
\put(1925,9170){\makebox(0,0)[r]{\strut{} 1000}}%
\put(2200,3300){\makebox(0,0){\strut{} 0}}%
\put(3568,3300){\makebox(0,0){\strut{} 100}}%
\put(4936,3300){\makebox(0,0){\strut{} 200}}%
\put(6304,3300){\makebox(0,0){\strut{} 300}}%
\put(7671,3300){\makebox(0,0){\strut{} 400}}%
\put(9039,3300){\makebox(0,0){\strut{} 500}}%
\put(10407,3300){\makebox(0,0){\strut{} 600}}%
\put(11775,3300){\makebox(0,0){\strut{} 700}}%
\put(687,6510){\rotatebox{90}{\makebox(0,0){\strut{}CPU time}}}%
\put(7262,2475){\makebox(0,0){\strut{}instances}}%
\put(4238,1375){\makebox(0,0)[r]{\strut{}maxsatz}}%
\put(4238,825){\makebox(0,0)[r]{\strut{}pbo}}%
\put(4238,275){\makebox(0,0)[r]{\strut{}msu1}}%
\put(9988,1375){\makebox(0,0)[r]{\strut{}msu2}}%
\put(9988,825){\makebox(0,0)[r]{\strut{}msu3}}%
\end{picture}%
\endgroup
\end{center}
    \caption{Run times on all instances}
    \label{fig:all-all}
\end{figure}

Figure~\ref{fig:all-all} plots the run times of each solver sorted by
increasing run times. As can be observed, the performance difference
for the \mxsat solvers considered is significant. {\tt msu2} and {\tt
  msu3} solve many more problem instances than any of the other
solvers.
As can also be observed in Figure~\ref{fig:all-all}, {\tt msu1}
exhibits a sharp transition between instances it can solve and
instances it is unable to solve. The reason is due to the size of the
computed unsatisfiable cores. For the more complex instances, the size
of the cores is significant, and so {\tt msu1} most often aborts due
to excessive memory requirements.

\begin{table}[t]
\begin{center}
\caption{Number of aborted instances, out of a total of 698 instances}
\label{tab:ab}
\begin{tabular}{|l|ccccc|}\hline
\mxsat solver     &  {\tt maxsatz}  &  PBO  &  {\tt msu1}  &  {\tt msu2}  &
{\tt msu3} \\ \hline
Aborted instances &  564            &  261  &  228   &  84    &  82  \\ \hline
\end{tabular}
\end{center}
\end{table}

A summary of the number of aborted instances is shown in
Table~\ref{tab:ab}. Over all instances, {\tt msu2} aborts 2 more
instances than {\tt msu3} (respectively 84 vs.~82), and both abort
significantly less instances than any of the other solvers. msu1
aborts 146 more instances than {\tt msu3} and 144 more instances than
{\tt msu2}. Somewhat surprisingly, the PBO model performs reasonably
well when compared with {\tt msu1}. As might be expected, {\tt
  maxsatz} aborts most industrial instances.

\begin{figure}[t]
  \hspace*{1cm}
  \begin{minipage}[t]{0.4\textwidth}
    \centerline{\scalebox{0.9}{
\begin{picture}(0,0)%
\includegraphics{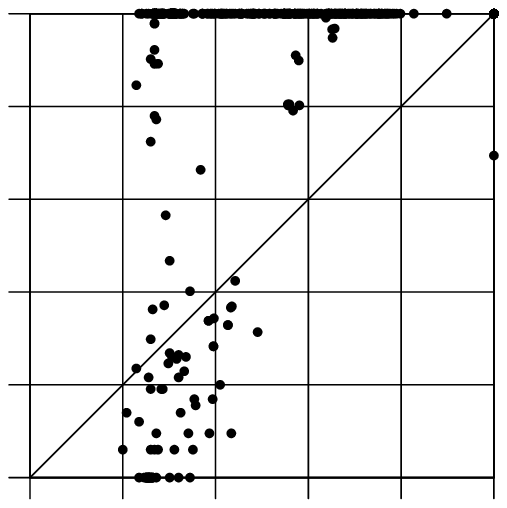}%
\end{picture}%
\begingroup
\setlength{\unitlength}{0.0200bp}%
\begin{picture}(15300,9180)(0,0)%
\put(3025,1950){\makebox(0,0)[r]{\strut{}$10^{-2}$}}%
\put(3025,3286){\makebox(0,0)[r]{\strut{}$10^{-1}$}}%
\put(3025,4622){\makebox(0,0)[r]{\strut{}$10^{0}$}}%
\put(3025,5958){\makebox(0,0)[r]{\strut{}$10^{1}$}}%
\put(3025,7294){\makebox(0,0)[r]{\strut{}$10^{2}$}}%
\put(3025,8630){\makebox(0,0)[r]{\strut{}$10^{3}$}}%
\put(3600,1100){\makebox(0,0){\strut{}$10^{-2}$}}%
\put(4936,1100){\makebox(0,0){\strut{}$10^{-1}$}}%
\put(6272,1100){\makebox(0,0){\strut{}$10^{0}$}}%
\put(7608,1100){\makebox(0,0){\strut{}$10^{1}$}}%
\put(8944,1100){\makebox(0,0){\strut{}$10^{2}$}}%
\put(10280,1100){\makebox(0,0){\strut{}$10^{3}$}}%
\put(1650,5290){\rotatebox{90}{\makebox(0,0){\strut{}maxsatz}}}%
\put(7215,275){\makebox(0,0){\strut{}msu1}}%
\end{picture}%
\endgroup
}}
  \end{minipage}
  \hfill
  \begin{minipage}[t]{0.4\textwidth}
    \centerline{\scalebox{0.9}{
\begin{picture}(0,0)%
\includegraphics{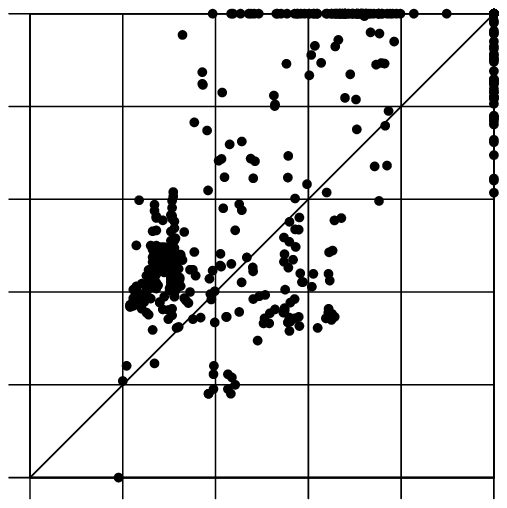}%
\end{picture}%
\begingroup
\setlength{\unitlength}{0.0200bp}%
\begin{picture}(15300,9180)(0,0)%
\put(3025,1950){\makebox(0,0)[r]{\strut{}$10^{-2}$}}%
\put(3025,3286){\makebox(0,0)[r]{\strut{}$10^{-1}$}}%
\put(3025,4622){\makebox(0,0)[r]{\strut{}$10^{0}$}}%
\put(3025,5958){\makebox(0,0)[r]{\strut{}$10^{1}$}}%
\put(3025,7294){\makebox(0,0)[r]{\strut{}$10^{2}$}}%
\put(3025,8630){\makebox(0,0)[r]{\strut{}$10^{3}$}}%
\put(3600,1100){\makebox(0,0){\strut{}$10^{-2}$}}%
\put(4936,1100){\makebox(0,0){\strut{}$10^{-1}$}}%
\put(6272,1100){\makebox(0,0){\strut{}$10^{0}$}}%
\put(7608,1100){\makebox(0,0){\strut{}$10^{1}$}}%
\put(8944,1100){\makebox(0,0){\strut{}$10^{2}$}}%
\put(10280,1100){\makebox(0,0){\strut{}$10^{3}$}}%
\put(1650,5290){\rotatebox{90}{\makebox(0,0){\strut{}pbo}}}%
\put(7215,275){\makebox(0,0){\strut{}msu1}}%
\end{picture}%
\endgroup
}}
  \end{minipage}
  \caption{Scatter plots on all instances maxsatz and PBO vs.~{\tt msu1}}
  \label{fig:scat-mxzpbo-msu1}
\end{figure}

Figure~\ref{fig:scat-mxzpbo-msu1} compares {\tt msu1} with {\tt
  maxsatz} and PBO, respectively. As can be seen, with the exception
of a few outliers, always taking negligible CPU time, {\tt msu1}
performs  significantly better than {\tt maxsatz}. In contrast, {\tt
  msu1} does not clearly outperform PBO. This is in part explained by
the quadratic encoding of AtMost~1 constraints in {\tt msu1}, which
causes a significant number of instances to abort. In addition, the
PBO model uses the most recent version of {\tt minisat+}, whereas the 
unsatisfiable core extractor used in all versions of the {\tt msu}
algorithm is based on {\tt minisat}~1.14.

\begin{figure}[t]
  \hspace*{1cm}
  \begin{minipage}[t]{0.4\textwidth}
    \centerline{\scalebox{0.9}{
\begin{picture}(0,0)%
\includegraphics{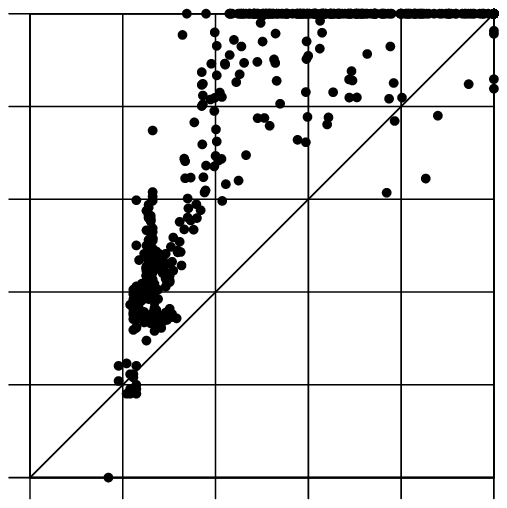}%
\end{picture}%
\begingroup
\setlength{\unitlength}{0.0200bp}%
\begin{picture}(15300,9180)(0,0)%
\put(3025,1950){\makebox(0,0)[r]{\strut{}$10^{-2}$}}%
\put(3025,3286){\makebox(0,0)[r]{\strut{}$10^{-1}$}}%
\put(3025,4622){\makebox(0,0)[r]{\strut{}$10^{0}$}}%
\put(3025,5958){\makebox(0,0)[r]{\strut{}$10^{1}$}}%
\put(3025,7294){\makebox(0,0)[r]{\strut{}$10^{2}$}}%
\put(3025,8630){\makebox(0,0)[r]{\strut{}$10^{3}$}}%
\put(3600,1100){\makebox(0,0){\strut{}$10^{-2}$}}%
\put(4936,1100){\makebox(0,0){\strut{}$10^{-1}$}}%
\put(6272,1100){\makebox(0,0){\strut{}$10^{0}$}}%
\put(7608,1100){\makebox(0,0){\strut{}$10^{1}$}}%
\put(8944,1100){\makebox(0,0){\strut{}$10^{2}$}}%
\put(10280,1100){\makebox(0,0){\strut{}$10^{3}$}}%
\put(1650,5290){\rotatebox{90}{\makebox(0,0){\strut{}pbo}}}%
\put(7215,275){\makebox(0,0){\strut{}msu2}}%
\end{picture}%
\endgroup
}}
  \end{minipage}
  \hfill
  \begin{minipage}[t]{0.4\textwidth}
    \centerline{\scalebox{0.9}{
\begin{picture}(0,0)%
\includegraphics{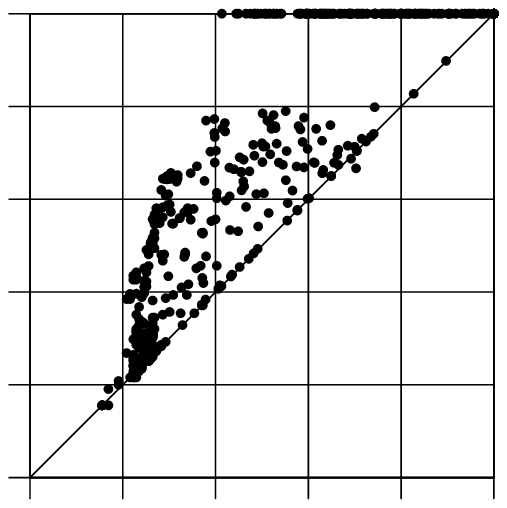}%
\end{picture}%
\begingroup
\setlength{\unitlength}{0.0200bp}%
\begin{picture}(15300,9180)(0,0)%
\put(3025,1950){\makebox(0,0)[r]{\strut{}$10^{-2}$}}%
\put(3025,3286){\makebox(0,0)[r]{\strut{}$10^{-1}$}}%
\put(3025,4622){\makebox(0,0)[r]{\strut{}$10^{0}$}}%
\put(3025,5958){\makebox(0,0)[r]{\strut{}$10^{1}$}}%
\put(3025,7294){\makebox(0,0)[r]{\strut{}$10^{2}$}}%
\put(3025,8630){\makebox(0,0)[r]{\strut{}$10^{3}$}}%
\put(3600,1100){\makebox(0,0){\strut{}$10^{-2}$}}%
\put(4936,1100){\makebox(0,0){\strut{}$10^{-1}$}}%
\put(6272,1100){\makebox(0,0){\strut{}$10^{0}$}}%
\put(7608,1100){\makebox(0,0){\strut{}$10^{1}$}}%
\put(8944,1100){\makebox(0,0){\strut{}$10^{2}$}}%
\put(10280,1100){\makebox(0,0){\strut{}$10^{3}$}}%
\put(1650,5290){\rotatebox{90}{\makebox(0,0){\strut{}msu1}}}%
\put(7215,275){\makebox(0,0){\strut{}msu2}}%
\end{picture}%
\endgroup
}}
  \end{minipage}
  \caption{Scatter plots on all instances, PBO and msu1 vs.~msu2}
  \label{fig:scat-pbomsu1-msu2}
\end{figure}

Figure~\ref{fig:scat-pbomsu1-msu2} compares {\tt msu2} against PBO and
{\tt msu1}. The performance difference is clear. {\tt msu2}
outperforms {\tt msu1} in almost all problem instances. For a very
small number of examples, {\tt msu1} can outperform {\tt msu2}, but
the differences are essentially negligible, never exceeding a small
percentage of the total run time. {\tt msu2} also clearly outperforms
PBO, aborting a fraction (close to 20\%) of the instances aborted by
PBO. However, a few outliers exist, and these are explained by the
fact that PBO uses the most recent version of {\tt minisat+}, and {\tt
  msu2} uses an unsatisfiable core extractor based on {\tt
  minisat}~1.14.

\begin{figure}[t]
  \hspace*{1cm}
  \begin{minipage}[t]{0.4\textwidth}
    \centerline{\scalebox{0.9}{
\begin{picture}(0,0)%
\includegraphics{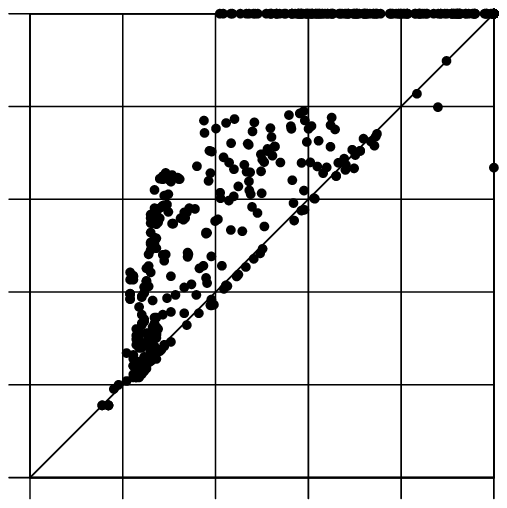}%
\end{picture}%
\begingroup
\setlength{\unitlength}{0.0200bp}%
\begin{picture}(15300,9180)(0,0)%
\put(3025,1950){\makebox(0,0)[r]{\strut{}$10^{-2}$}}%
\put(3025,3286){\makebox(0,0)[r]{\strut{}$10^{-1}$}}%
\put(3025,4622){\makebox(0,0)[r]{\strut{}$10^{0}$}}%
\put(3025,5958){\makebox(0,0)[r]{\strut{}$10^{1}$}}%
\put(3025,7294){\makebox(0,0)[r]{\strut{}$10^{2}$}}%
\put(3025,8630){\makebox(0,0)[r]{\strut{}$10^{3}$}}%
\put(3600,1100){\makebox(0,0){\strut{}$10^{-2}$}}%
\put(4936,1100){\makebox(0,0){\strut{}$10^{-1}$}}%
\put(6272,1100){\makebox(0,0){\strut{}$10^{0}$}}%
\put(7608,1100){\makebox(0,0){\strut{}$10^{1}$}}%
\put(8944,1100){\makebox(0,0){\strut{}$10^{2}$}}%
\put(10280,1100){\makebox(0,0){\strut{}$10^{3}$}}%
\put(1650,5290){\rotatebox{90}{\makebox(0,0){\strut{}msu1}}}%
\put(7215,275){\makebox(0,0){\strut{}msu3}}%
\end{picture}%
\endgroup
}}
  \end{minipage}
  \hfill
  \begin{minipage}[t]{0.4\textwidth}
    \centerline{\scalebox{0.9}{
\begin{picture}(0,0)%
\includegraphics{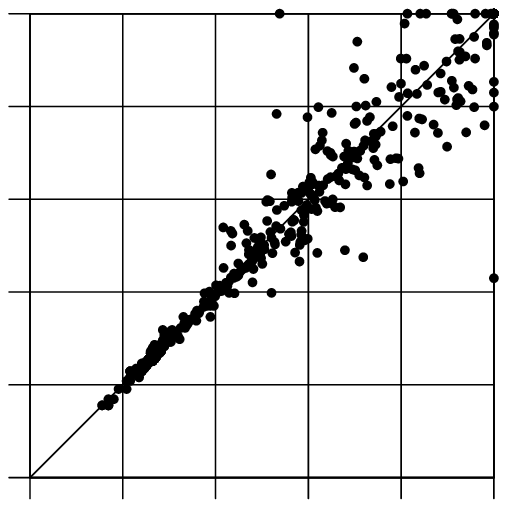}%
\end{picture}%
\begingroup
\setlength{\unitlength}{0.0200bp}%
\begin{picture}(15300,9180)(0,0)%
\put(3025,1950){\makebox(0,0)[r]{\strut{}$10^{-2}$}}%
\put(3025,3286){\makebox(0,0)[r]{\strut{}$10^{-1}$}}%
\put(3025,4622){\makebox(0,0)[r]{\strut{}$10^{0}$}}%
\put(3025,5958){\makebox(0,0)[r]{\strut{}$10^{1}$}}%
\put(3025,7294){\makebox(0,0)[r]{\strut{}$10^{2}$}}%
\put(3025,8630){\makebox(0,0)[r]{\strut{}$10^{3}$}}%
\put(3600,1100){\makebox(0,0){\strut{}$10^{-2}$}}%
\put(4936,1100){\makebox(0,0){\strut{}$10^{-1}$}}%
\put(6272,1100){\makebox(0,0){\strut{}$10^{0}$}}%
\put(7608,1100){\makebox(0,0){\strut{}$10^{1}$}}%
\put(8944,1100){\makebox(0,0){\strut{}$10^{2}$}}%
\put(10280,1100){\makebox(0,0){\strut{}$10^{3}$}}%
\put(1650,5290){\rotatebox{90}{\makebox(0,0){\strut{}msu2}}}%
\put(7215,275){\makebox(0,0){\strut{}msu3}}%
\end{picture}%
\endgroup
}}
  \end{minipage}
  \caption{Scatter plots on all instances, msu1 and msu2 vs.~msu3}
  \label{fig:scat-msuX-msu3}
\end{figure}

Figure~\ref{fig:scat-msuX-msu3} compares {\tt msu3} against {\tt msu1}
and {\tt msu2}. {\tt msu1} performs significantly worse than {\tt
  msu3}, with a few outliers, only one of which is relevant. In
contrast, {\tt msu2} and {\tt msu3} perform similarly, even though
{\tt msu2} usually exhibits smaller run times.
Nevertheless, the results also suggest that {\tt msu3} can be an
interesting alternative to {\tt msu2} for a reasonable number of
problem instances.

The previous results provide clear evidence that
unsatisfiability-based \mxsat algorithms are effective for solving 
problem instances obtained from industrial settings. However, several
of the problem instances considered, albeit unsatisfiable, do not
represent problems originally formulated as \mxsat problems.
Recent work has shown that \mxsat has practical application in FPGA
routing~\cite{kas-tcad03} and system debugging~\cite{kas-fmcad07}.
However, and motivated by the limitations of existing \mxsat solvers,
these \mxsat instances were solved with approximate algorithms.
Our results indicate that unsatisfiability-based \mxsat algorithms are
very efficient at solving problem instances from design debugging, but 
less effective at solving FPGA routing instances.
The class FPGA contains 16 unsatisfiable instances. Of these 16, {\tt
  msu2} solves 3, {\tt msu1} solves 2, {\tt msu3} solves 1, and {\tt
  maxsatz} solves 1.
It is well-known that SAT instances from FPGA routing have specific
structure, that makes them quite difficult for SAT
solvers~\cite{kas-fpga99,kas-tcad03}. This in part explains the
results of all \mxsat solvers on these instances.

\begin{table}[t]
\label{tab:dd:res}
\caption{Results for design debugging instances}
\begin{center}
\begin{tabular}{|l|rrr|rrrrr|} \hline
Instance & \#V & \#C & solution & maxsatz & pbo & msu1 & msu2 & msu3 \\ \hline
b14\_opt\_bug2\_vec1-gt-0 & 130328 & 402707 & 402706 & -- & -- & 10.12 & {\bf 9.79} & 11.74 \\ \hline
b15-bug-4vec-gt-0 & 581064 & 1712690 & -- & -- & -- & -- & -- & -- \\ \hline
b15-bug-1vec-gt-0 & 121836 & 359040 & 359039 & -- & -- & -- & {\bf 27.43} & 93.47 \\ \hline
c1\_DD\_s3\_f1\_e2\_v1-bug-4vec-gt-0 & 391897 & 989885 & 989881 & -- & -- & {\bf 41.90} & {\bf 41.90} & 47.29 \\ \hline
c1\_DD\_s3\_f1\_e2\_v1-bug-1vec-gt-0 & 102234 & 258294 & 258293 & -- & -- & {\bf 5.89} & 5.92 & 7.04 \\ \hline
c2\_DD\_s3\_f1\_e2\_v1-bug-4vec-gt-0 & 400085 & 1121810 & 1121806 & -- & -- & 98.25 & {\bf 51.89} & 249.42 \\ \hline
c2\_DD\_s3\_f1\_e2\_v1-bug-1vec-gt-0 & 84525 & 236942 & 236941 & -- & -- & 16.03 & {\bf 5.72} & 6.72 \\ \hline
c3\_DD\_s3\_f1\_e1\_v1-bug-4vec-gt-0 & 33540 & 86944 & 86940 & -- & -- & 2.92 & {\bf 2.84} & 3.21 \\ \hline
c3\_DD\_s3\_f1\_e1\_v1-bug-1vec-gt-0 & 8385 & 21736 & 21735 & -- & 590.12 & {\bf 0.44} & {\bf 0.44} & 0.49 \\ \hline
c4\_DD\_s3\_f1\_e1\_v1-bug-gt-0 & 797728 & 2011216 & 2011208 & -- & -- & 137.19 & {\bf 136.72} & 148.02 \\ \hline
c4\_DD\_s3\_f1\_e2\_v1-bug-4vec-gt-0 & 448465 & 1130672 & 1130668 & -- & -- & 47.23 & {\bf 47.22} & 53.20 \\ \hline
c4\_DD\_s3\_f1\_e2\_v1-bug-1vec-gt-0 & 131584 & 331754 & 331753 & -- & -- & 7.70 & {\bf 7.62} & 9.06 \\ \hline
c5315-bug-gt-0 & 1880 & 5049 & 5048 & 169.44 & 3.18 & {\bf 0.14} & {\bf 0.14} & 0.15 \\ \hline
c5\_DD\_s3\_f1\_e1\_v1-bug-4vec-gt-0 & 100472 & 270492 & 270488 & -- & -- & 10.26 & {\bf 10.20} & 11.49 \\ \hline
c5\_DD\_s3\_f1\_e1\_v1-bug-gt-0 & 200944 & 540984 & 540976 & -- & -- & {\bf 33.26} & 33.32 & 36.25 \\ \hline
c5\_DD\_s3\_f1\_e1\_v1-bug-1vec-gt-0 & 25118 & 67623 & 67622 & -- & -- & 1.47 & {\bf 1.46} & 1.69 \\ \hline
c5\_DD\_s3\_f1\_e1\_v2-bug-gt-0 & 200944 & 540984 & 540976 & -- & -- & {\bf 33.22} & 33.32 & 36.17 \\ \hline
c6288-bug-gt-0 & 3462 & 9285 & 9284 & -- & 17.12 & 6.01 & 0.54 & {\bf 0.45} \\ \hline
c6\_DD\_s3\_f1\_e1\_v1-bug-4vec-gt-0 & 170019 & 454050 & 454046 & -- & -- & 17.79 & {\bf 17.63} & 19.91 \\ \hline
c6\_DD\_s3\_f1\_e1\_v1-bug-gt-0 & 298058 & 795900 & 795892 & -- & -- & 50.71 & {\bf 50.69} & 54.98 \\ \hline
c6\_DD\_s3\_f1\_e1\_v1-bug-1vec-gt-0 & 44079 & 117720 & 117719 & -- & -- & 2.61 & {\bf 2.57} & 3.03 \\ \hline
c6\_DD\_s3\_f1\_e2\_v1-bug-4vec-gt-0 & 170019 & 454050 & 454046 & -- & -- & 17.81 & {\bf 17.62} & 19.95 \\ \hline
c7552-bug-gt-0 & 2640 & 7008 & 7007 & -- & 55.06 & 0.81 & {\bf 0.21} & {\bf 0.21} \\ \hline
mot\_comb1.\_red-gt-0 & 2159 & 5326 & 5325 & 0.15 & 1.08 & {\bf 0.14} & {\bf 0.14} & 0.16 \\ \hline
mot\_comb2.\_red-gt-0 & 5484 & 13894 & 13893 & 6.71 & 2.58 & {\bf 0.29} & {\bf 0.29} & 0.33 \\ \hline
mot\_comb3.\_red-gt-0 & 11265 & 29520 & 29519 & -- & 67.43 & {\bf 0.59} & {\bf 0.59} & 0.66 \\ \hline
s15850-bug-4vec-gt-0 & 88544 & 206252 & 206248 & -- & -- & {\bf 7.55} & 7.63 & 8.42 \\ \hline
s15850-bug-1vec-gt-0 & 22136 & 51563 & 51562 & -- & 26.01 & 1.08 & {\bf 1.07} & 1.23 \\ \hline
s38584-bug-1vec-gt-0 & 314272 & 819830 & 819829 & -- & -- & 23.43 & {\bf 21.12} & 25.41 \\ \hline
\end{tabular}
\end{center}
\end{table}

For the design debugging instances the results are quite different.
Table~\ref{tab:dd:res} shows the CPU times for all \mxsat solvers on
all design debugging instances~\cite{kas-fmcad07}. As before, the
unsatisfiability-based \mxsat algorithms perform remarkably better
than the other algorithms, {\tt maxsatz} and the PBO model. In
addition, and also as before, {\tt msu2} is the best performing
algorithm, and aborts only one instance. {\tt msu3} also aborts only
one instance, but in general performs worse than {\tt msu2}. Finally,
{\tt msu1} aborts 2 instances, and performs worse than {\tt msu2} for
almost all instances. For instances with large unsatisfiable cores
(e.g.~b15-bug-onevec-gate-0) the linear encoding used in {\tt msu2} 
ensures manageable size representations of the AtMost~1 constraints.
The same holds true to {\tt msu3}, for small values of $k$. In
contrast, {\tt msu1} uses a quadratic encoding and so it often aborts
instances with large unsatisfiable cores.
Moreover, for instances requiring the identification of several
unsatisfiable cores sharing common clauses, the additional constraints
proposed in Section~\ref{ssec:bvconstr} are useful for {\tt msu2}. It
should be observed that the only design debugging instance that is
aborted by both {\tt msu2} and {\tt msu3} is also aborted by the
unsatisfiable core extractor, again  suggesting that performance of
unsatisfiability-based \mxsat solvers is indexed to the efficiency of
SAT solvers.


\section{Conclusions}
\label{sec:conc}

Recent work has shown that \mxsat has a number of significant
practical applications~\cite{kas-fmcad07}. However, current state of
the art \mxsat solvers are ineffective on most problem instances
obtained from practical applications.

This paper focus on solving \mxsat problem instances obtained form
practical applications, and conducts a detailed analysis of \mxsat
algorithms based on unsatisfiable subformula identification. Moreover,
the paper develops improvements to existing algorithms and proposes a
new \mxsat algorithm. The proposed improvements ({\tt msu2}) and new
algorithm ({\tt msu3}) provide significant performance improvements,
and allow indexing the hardness of solving practical instances of
\mxsat to the ability of modern SAT solvers for proving
unsatisfiability and identifying unsatisfiable subformulas. The
algorithms described in this paper are by far the most effective for
instances obtained from practical applications, clearly outperforming
existing state of the art \mxsat solvers, and further improvements are
to be expected.

Despite the promising results of the new \mxsat algorithms, a number
of research directions can be envisioned. As the experimental results 
show, the role of encoding cardinality constraints is significant, and
an extensive evaluation of alternative encodings should be
considered. 
The unsatisfiable core extractor is based on {\tt minisat}~1.14. A
core extractor based a more recent SAT solver is expected to
improve the efficiency of {\tt msu1}, {\tt msu2} and {\tt msu3}.
Finally, the problem instances for the FPGA routing problem are still
challenging, even though the new \mxsat algorithms can solve some of
these instances, and so motivate the development of further
improvements to \mxsat algorithms.


\vspace*{-0.15cm}
\subsubsection{\ackname}
This work is partially supported by EU project IST/033709 and by
EPSRC grant EP/E012973/1. 
\vspace*{-0.15cm}

\bibliographystyle{abbrv}

\end{document}